\documentclass[preprint,12pt,authoryear]{elsarticle}

\usepackage{amssymb}

\usepackage{algorithm}
\usepackage{algorithmic}

\usepackage{multirow}

\usepackage{pifont}

\usepackage{graphicx}
\usepackage{amsmath}
\usepackage{amssymb}
\usepackage{booktabs}

\usepackage{setspace}
\usepackage{hyperref}
\usepackage[hyphenbreaks]{breakurl}

\usepackage{geometry}
\usepackage[capitalize]{cleveref}
\crefname{section}{Sec.}{Secs.}
\Crefname{section}{Section}{Sections}
\Crefname{table}{Table}{Tables}
\crefname{table}{Tab.}{Tabs.}

\begin{document}

\begin{frontmatter}



\title{Adversarial Camera Patch: An Effective and Robust Physical-World Attack on Object Detectors}


\author[label1]{Kalibinuer Tiliwalidi}
\ead{ka202011081727@163.com}

\author[label1]{Ling Tian}
\ead{lingtian@uestc.edu.cn}


\author[label1]{Xu Zheng}
\ead{xzheng@uestc.edu.cn}

\author[label1]{Chengyin Hu}
\ead{cyhuuestc@gmail.com}

\affiliation[label1]{organization={School of Computer Science and Engineering, University of Electronic Science and Technology of China},
            addressline={No. 2006, Xiyuan Avenue, Gaoxin District},
            city={Chengdu},
            postcode={611731},
            state={Sichuan},
            country={China}}

\cortext[cor1]{Corresponding author}

\begin{abstract}
Nowadays, the susceptibility of deep neural networks (DNNs) has garnered significant attention. Researchers are exploring patch-based physical attacks, yet traditional approaches, while effective, often result in conspicuous patches covering target objects. This leads to easy detection by human observers. Recently, novel camera-based physical attacks have emerged, leveraging camera patches to execute stealthy attacks. These methods circumvent target object modifications by introducing perturbations directly to the camera lens, achieving a notable breakthrough in stealthiness.
However, prevailing camera-based strategies necessitate the deployment of multiple patches on the camera lens, which introduces complexity. To address this issue, we propose Adversarial Camera Patch (\textbf{ADCP}). ADCP employs a single camera patch, optimizing its physical parameters using Particle Swarm Optimization (PSO) to achieve maximal adversarial impact. With these parameters, the camera patch is applied to the lens, generating effective physical samples. Our method's efficacy is validated through ablation experiments in a digital setting, consistently demonstrating strong adversarial impact even in worst-case scenarios. Notably, ADCP exhibits higher robustness compared to baselines in both digital and physical domains. Employing the generated samples for advanced object detection attack, we showcase robust transfer attack capability.
Given its simplicity, robustness, and stealthiness, we advocate for the attention and consideration of the ADCP framework as it presents an avenue for achieving streamlined, potent, and stealthy attacks.

\end{abstract}




\end{frontmatter}



\section{Introduction}

\begin{figure}
\centering
\includegraphics[width=1\columnwidth]{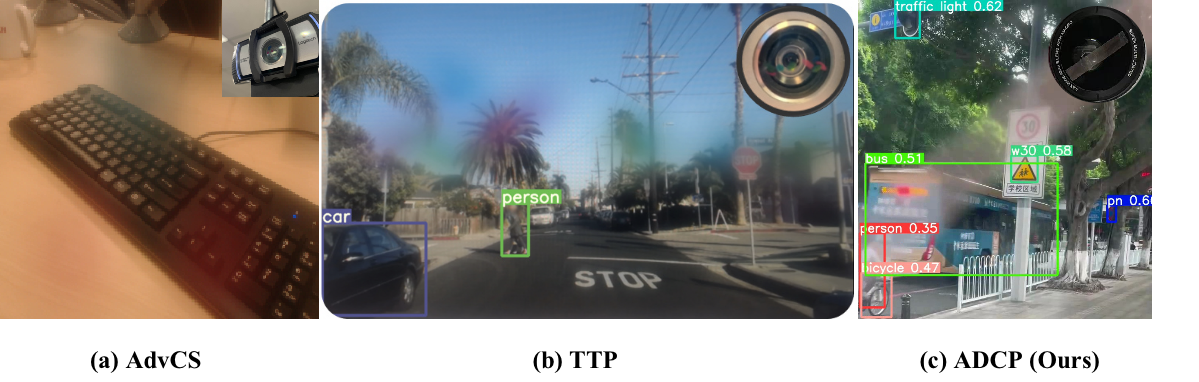} 
\caption{Demonstration of our proposed ADCP and other camera-based attacks.}.
\label{figure1}
\end{figure}

In the realm of computer vision systems' security and dependability, the emergence of physical attacks \cite{ref1} has garnered escalating interest as a significant threat. The extensive proliferation of computer vision technology has yielded remarkable breakthroughs across diverse domains, including autonomous driving, facial recognition, and security surveillance. However, this widespread integration has concurrently unveiled a susceptibility to physical attacks. Unlike conventional digital attacks, physical attacks capitalize on real-world physical attributes such as light, sound, and temperature to manipulate, disrupt, or undermine the normal operations of computer vision systems. These attacks can precipitate erroneous system interpretations, erroneous decision-making, security vulnerabilities, and in the gravest scenarios, precipitate substantial security risks.

At present, a significant portion of physical attacks \cite{ref2,ref3,ref4} utilizes adversarial patches as perturbations to execute physical attacks against advanced object detectors. Adversarial patches commonly cover a substantial fraction of the target object's area, bolstering the robustness of physical attacks at the expense of reduced stealthiness. While patch-based physical attacks maintain the semantic integrity of the target object, the conspicuousness of the perturbation remains a challenge. To address this issue, some studies have introduced light-based physical attacks (e.g., lasers \cite{ref5}, projectors \cite{ref6}). These leverage the transitory nature of illumination to instantaneously project an optimized beam onto the target object's surface, facilitating immediate physical attacks. Differing from patch-based counterparts, light-based methods can manipulate the light source's on-off state, projecting the beam during attacks and extinguishing the source when dormant. This results in superior stealthiness for light-based methods, as the physical perturbation isn't persistently affixed to the target object's surface. While light-based physical attacks enhance stealthiness, they often come at the cost of reduced robustness. Similarly, both light-based and patch-based physical attacks share a commonality: they entail modifications to the target object. To tackle this limitation, certain studies have proposed camera-based physical attacks \cite{ref7,ref8}. These execute covert physical attacks by attaching inconspicuous patches directly onto the camera lens. Such camera-based methods execute attacks by modifying the camera itself instead of altering the target object, thus achieving an elevated level of stealthiness. However, prevailing camera-based strategies involve deploying numerous small patches onto the lens, leading to unwanted errors. Additionally, the intricate process of patch deployment onto the lens presents challenges in practical implementation.

\begin{table} 
	\centering
    \setlength{\belowcaptionskip}{10pt}
    \caption{The comparison between existing methods and our method.}.
    \label{Table1}
	\begin{tabular}{cccc}

    \hline
    Method & Number of patches & Scenario & Manufacturing time\\
    \hline
    AdvCS \cite{ref7} & 6 & White-box & Around 2 hours\\
    \hline
    TTP \cite{ref8} & 8 & White-box & Around 2 hours\\
    \hline
    ADCP (Ours) & 1 & Black-box & Around 0.5 hours\\
    \hline

\end{tabular}
\vspace{-0.2cm}
\end{table}

In this study, we introduce a pioneering camera-based attack named adversarial camera patch. Figure \ref{figure1} vividly illustrates the deployment contrasts between our proposed approach and the baseline strategy. Drawing from the preceding discourse and Figure \ref{figure1}, it becomes apparent that our method holds a distinct advantage over prevailing camera-based physical attacks. We employ a singular translucent patch for executing physical attacks, a design that not only mitigates potential experimental errors stemming from perturbation deployment but also streamlines the operational complexities, rendering our method more practically viable. To further elucidate the differentiation between our method and the baseline, we showcase the comparison in Table \ref{Table1}. It emerges that our approach entails fewer perturbations and is notably facile to deploy. Enacting the proposed attack unfolds in several steps. Firstly, the physical parameters of the camera patch are precisely formulated. These encompass four vital parameters—position, width, transparency, and color—utilized to generate the simulated camera patch. Subsequently, employing a straightforward linear synthesis technique, we amalgamate these simulated camera patches with pristine images to generate digital samples. Thereafter, the Particle Swarm Optimization (PSO) algorithm \cite{ref9} is harnessed to optimize the physical parameters of the camera patch, culminating in the creation of the most adversarial digital samples. Ultimately, the physical camera patch is affixed to the camera lens, capturing physical samples for the purpose of executing the physical attacks.

\begin{figure}
\centering
\includegraphics[width=1\columnwidth]{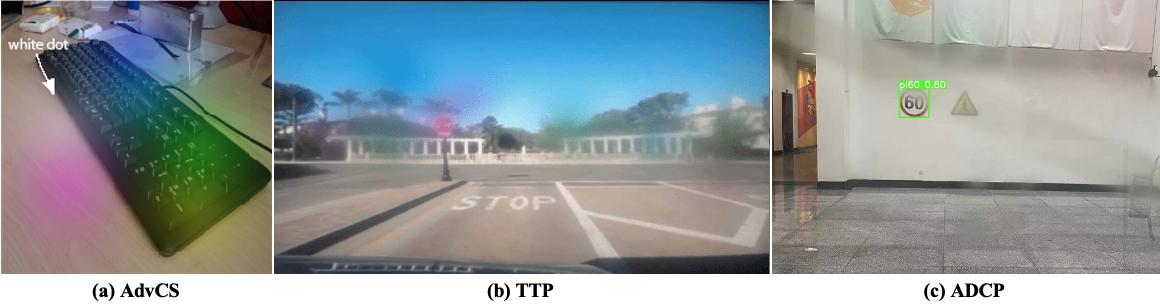} 
\caption{Visual comparison.}.
\label{figure2}
\end{figure}

Our approach offers simplicity in both optimization and deployment in the physical realm. Following the acquisition of simulation samples via PSO optimization, the application of a camera patch to the camera lens facilitates the execution of flexible and inconspicuous physical attacks. The robustness and stealthiness of our method are noteworthy. In terms of robustness, our method achieves an attack success rate of 96.31\% (pl60+w50) and 78.16\% (pl60+pl60) during indoor testing, and an impressive 88.31\% during outdoor testing. Concerning stealthiness, Figure \ref{figure2} presents a comparative analysis between our method and the physical samples engendered by baseline approaches. It is evident that our method demonstrates superior stealthiness, with the presence of perturbation remaining largely imperceptible without meticulous scrutiny. Furthermore, when compared with the existing camera-based techniques, our method introduces subtler perturbations, thereby enhancing its concealment in comparison to baseline methodologies. The requisite physical devices for our method are both economical and widely accessible, with the collective cost of the experimental apparatus amounting to less than \$7. This financial accessibility renders our approach feasible for deployment by a broad spectrum of potential attackers.

Our contributions can be outlined as follows:

\begin{itemize}

\item Novel Adversarial Camera-Based Physical Attack (ADCP): We introduce ADCP, a novel black-box camera-based method for robustly generating adversarial examples to execute physical attacks on advanced object detectors. Our approach leverages PSO optimization, resulting in the creation of robust adversarial perturbations. Notably, ADCP surpasses baselines in terms of robustness, stealthiness, and deployment simplicity. The economical nature of our method ensures it can be practically implemented at a cost not exceeding \$7, rendering it readily applicable in real-world scenarios.

\item Comparative Analysis and Method Advantages: Through a comprehensive evaluation, we conduct a meticulous comparison of existing physical attack techniques. This analysis underscores the distinctive benefits of our approach in relation to its counterparts. By substantiating our findings with exhaustive experiments, we establish the robustness of ADCP across various distance angles, all while preserving its stealthy nature. These empirical results affirm the utility of ADCP as a noteworthy advancement within the domain of camera-based physical attacks.

\item Thorough Method Analysis: Our study extends to a comprehensive investigation of the proposed ADCP method. This includes ablation experiments that affirm the method's consistent performance across diverse color settings. Additionally, digital and physical transfer attack experiments substantiate the method's effectiveness in achieving superior transfer attack capabilities. These analyses contribute to our comprehensive understanding of the method's performance and its practical viability.

\end{itemize}

\section{Related works}

\subsection{Digital attacks}

The concept of adversarial attacks was initially introduced by \cite{ref10}, who demonstrated the susceptibility of sophisticated deep neural networks (DNNs) to minor perturbations. This seminal work laid the foundation for subsequent developments in the field, leading to the emergence of an array of digital attack techniques \cite{ref11,ref12,ref13,ref14}.

Digital attacks, by and large, necessitate perturbations that remain imperceptible to the human eye yet ensure their effectiveness. Among various strategies, the use of ${l}_{2}$ and ${l}_{\infty}$ norms has emerged as common practice to constrain the magnitude of perturbations \cite{ref15,ref16}. Researchers have also explored alternative avenues for executing digital attacks, involving the modification of different attributes within images. Notably, endeavors \cite{ref17,ref18,ref19} have introduced the manipulation of colors to orchestrate adversarial attacks within digital environments. The resultant adversarial samples can induce local color changes while preserving semantic integrity, thereby effectively deceiving advanced DNNs. Other approaches \cite{ref3,ref4,ref20,ref23} generate adversarial samples by overlaying textures or camouflage upon the target object's surface. Although these textures and disguises frequently envelop substantial portions or the entirety of the target object, their conspicuousness often renders them easily discernible by human observers. Nonetheless, based on the target's visual characteristics, these perturbations can be regarded as plausible, even if they completely obscure the target object. Certain researchers have explored the alteration of digital image's physical parameters to generate adversarial samples \cite{ref24,ref25}. By selectively modifying local segments of the target object through rendering, only the essential content of the object is retained. The resulting adversarial examples introduce subtle distortions that can effectively mislead advanced DNNs. More recently, scholars have introduced digital attacks grounded in natural phenomena \cite{ref26,ref27}. Through simulating the visual effects of raindrops, artificial raindrop patterns are superimposed onto clean samples, thereby generating natural-looking adversarial digital samples. Importantly, these perturbed samples evade detection due to their congruence with natural scene variations, and the inclusion of raindrops is deemed reasonable within such contexts.

It is imperative to highlight that the foundational assumption underpinning digital adversarial attacks is the attacker's unfettered ability to directly manipulate the input image—an assumption that lacks practicality in the realm of physical attacks. Unlike their digital counterparts, physical attacks necessitate a more elaborate sequence of steps. Primarily, the physical attack involves capturing an image of the target object using a camera. Subsequently, this acquired image, now imbued with added perturbations, is presented to the target model to initiate and execute the attack.

\subsection{Physical attacks}

\textbf{Patch-based Attacks:} Patch-based attacks, involve the strategic use of meticulously crafted patches affixed to the target object's surface to deceive advanced Deep Neural Networks \cite{ref28}. Generally, these attacks entail the application of patches that cover substantial portions, if not the entirety, of the target object's area, with relatively less regard for perturbation constraints \cite{ref29}. Although these patches extensively envelop the target object, its inherent semantic information remains largely unaltered. Notably, recent research has focused on striking a balance between the patch's stealthiness and its impact on attack efficacy. \cite{ref30} pioneered the use of total variation loss to generate smoother adversarial patches that occupy a smaller portion of the target area, successfully compromising advanced pedestrian target detection systems. Their experimental findings corroborate the marked reduction in the accuracy of the pedestrian detector when subjected to the generated patches. Building upon this, \cite{ref31} devised a strategy wherein optimized adversarial patches were printed onto clothing to diminish their conspicuity to human observers. Their approach showcased its effectiveness across white-box and black-box settings. Similarly, \cite{ref32} introduced a method involving cartoon-like patches that hoodwink pedestrian detectors, resulting in patches that appear more organic and inconspicuous. Their innovative framework employs a two-stage training approach to generate adversarial patches with enhanced naturalness and rationality. The experiments underscored the substantial efficacy of this method across both simulated and physical attacks. In summation, patch-based attacks concentrate on the meticulous design of textures or patterns capable of deceiving target detectors, often with relatively less emphasis on other patch properties such as size and shape.

\textbf{Light-Based Attacks:} Light-based attacks involve the execution of instantaneous physical attacks against DNNs through the precise deployment of meticulously designed light beams projected onto the target object's surface \cite{ref33}. Leveraging the instantaneous characteristic of light beams, light-based attacks inherently exhibit superior stealthiness compared to patch-based attacks. \cite{ref34} introduced Adversarial Laser Beams (AdvLB), employing laser beams as the instrumental means for crafting and executing physical attacks against advanced DNNs. Through simulation and optimization of the physical attributes of laser beams, their methodology translated into real-world physical attacks using laser pointers. Their experimental findings underscored the physical efficacy of AdvLB. However, AdvLB is susceptible to spatial errors in real-world deployment, thereby limiting its reliability. Inspired by AdvLB, \cite{ref5} developed Adversarial Laser Points (AdvLS), wherein carefully calibrated laser points are projected onto the target object's surface for effective and discreet black-box physical attacks. AdvLS's operational window spans daytime, and the inconspicuousness of the perturbation patterns produced by laser points renders them inconspicuous to human observers. Yet, AdvLS is susceptible to interference from sunlight during daylight hours, impacting its performance. \cite{ref6} introduced the Optical Adversarial Attack (OPAD), wherein a projector is utilized as a perturbation light source to project precisely devised perturbations onto the target object's surface to confound advanced DNNs. OPAD magnifies imperceptible perturbations, allowing the projected patterns to be captured by cameras for subsequent attacks. However, OPAD is constrained to nocturnal operations. In summary, light-based attacks harness the characteristics of instantaneous action to achieve temporal stealthiness, setting them apart through their ability to deceive DNNs in a swift and transient manner.

\textbf{Natural-phenomenon-based Attacks:} Natural-phenomenon-based attacks involve the simulation of patterns derived from natural occurrences (e.g., shadows, raindrops, etc.) and their strategic deployment onto target objects to generate physical samples for subsequent attacks. \cite{ref35} introduced the shadow attack paradigm, utilizing shadows as perturbations. Employing simulation to achieve the shadow effect, the generated adversarial samples were further refined through evolutionary algorithms. A cardboard structure was subsequently employed to generate shadow disturbances in front of the target object. Experimental results validated the physical efficacy of the shadow attack methodology. However, the shadow attack's performance becomes constrained under conditions of low ambient lighting. \cite{ref36} delved into the realm of raindrop-based physical attacks, introducing AdvRain. By strategically deploying 20 raindrops onto the target object's surface, an effective physical attack on the target model was achieved. AdvRain executed covert physical attacks, although its robustness remained uncertain. In sum, natural-phenomenon-based attacks leverage inherent occurrences from the natural world to execute surreptitious physical attacks, representing a relatively novel avenue within the landscape of physical attacks.

\textbf{Camera-based Attacks:} Camera-based attacks involve the application of an adversarial patch onto a camera lens, devoid of any alteration to the target object. This approach captures a physical sample of the target object through the camera, subsequently deceiving DNNs. Distinguished by its nonintrusive character, camera-based attacks confer a stealth advantage compared to patch-based, light-based, and natural-phenomenon-based attacks, which necessitate target object modifications. \cite{ref7} introduced the AdvCS camera-based attack by lens modification. Small patches were affixed to the camera lens, manifesting as color perturbations upon image capture. Experimental validations confirmed the efficacy of this camera-based physical attacks. Notably, AdvCS perturbations were non-transparent, accentuating sample conspicuousness. Additionally, AdvCS targeted classifiers rather than detectors, an uncommon scenario in practical applications. Addressing these limitations, \cite{ref8} proposed TTP, a refined camera-based physical attack. Employing translucent patches as perturbations improved sample concealment, culminating in a 42.27\% attack success rate against advanced detectors. Yet, the subtlety of TTP came at the expense of attack robustness. In essence, prevailing camera-based methods confer spatial invisibility without affecting target objects. Nevertheless, AdvCS and TTP's reliance on multiple small patches for lens coverage culminated in operational complexity and substantial experimental errors, limiting attack performance and feasibility. Contrasting these approaches, our method circumvents these challenges through the application of a solitary translucent patch as perturbation.

\section{Method}

In this study, akin to the explanation provided by \cite{ref8}, we postulate that the attacker possesses direct access to the camera lens, enabling the deployment of meticulously optimized perturbations upon its surface.

\subsection{Problem definition}

Considering a dataset $D$ comprising clean samples, we delineate two distinct sets, $X$ and $Y$, signifying the aggregation of clean samples and the assemblage of ground truth labels, correspondingly. With $f$ representing the pre-trained model of the object detector, for every $X$ belonging to the dataset $D$, the object detector $f: X \rightarrow Y$ adeptly prognosticates the label $y$ for the clean sample. Within $y$, three pivotal elements are encapsulated: ${V}_{pos}$, denoting the positional intel of the bounding box; ${V}_{obj}$, encapsulating the confidence level of the target object; and ${V}_{cls}$, conveying the category of the prognosticated entity:

\begin{equation}
    \label{Formula 1}
    y:=[{V}_{pos},{V}_{obj},{V}_{cls}]=f(X)
\end{equation}

\begin{figure}
\centering
\includegraphics[width=1\columnwidth]{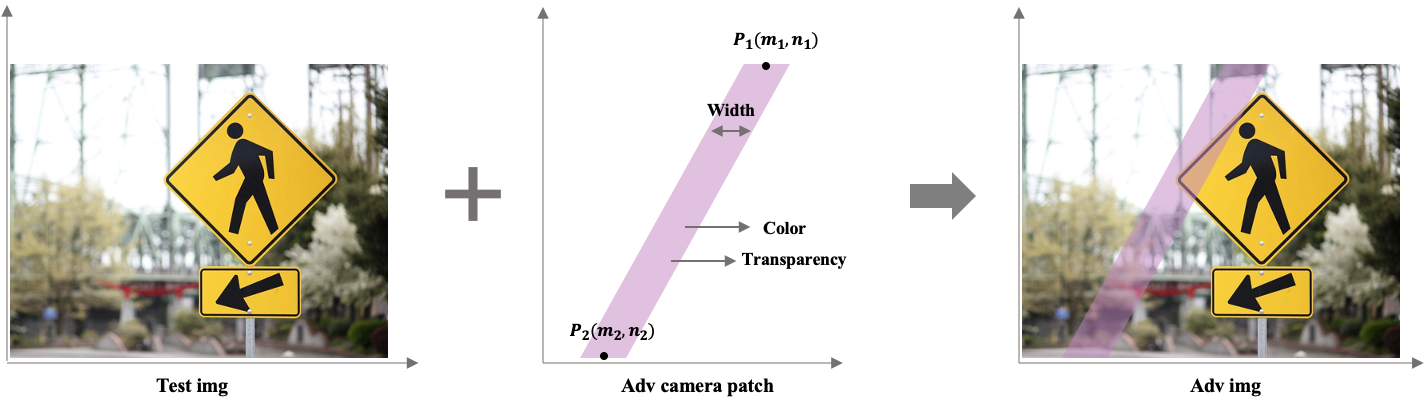} 
\caption{Camera patch modeling. ADCP exploits the physical parameters of the most adversarial camera patches to perform physical attacks, and it can be seen that our approach is simple to operate and flexible.}.
\label{figure3}
\end{figure}

\subsection{Camera patch modeling}

Our approach employs a translucent patch, affixed onto the camera lens, to execute a black-box physical attack. The camera patch is conceptually depicted in Figure \ref{figure3}, and its characteristics are represented by four distinct physical parameters: position, color, width, and transparency. In the ensuing discussion, each of these physical parameters will be expounded upon in detail.

\textbf{Position $PS$:} The position parameter, denoted as $PS$, is determined by the coordinates of the two endpoints of a line, effectively pinpointing the precise location of the camera patch within the image. Our methodology constrains these two endpoints to lie along the upper and lower edges of the image. Hence, the position $PS$ can be represented as a set comprising two points, i.e., $PS = \{{PS}_{1}({m}_{1}, {n}_{1}), {PS}_{2}({m}_{2}, {n}_{2})\}$. In our experimental setup, we keep ${n}_{1}$ and ${n}_{2}$ as constant values, ensuring that the camera patch's position information varies solely in the horizontal plane of the image. While, in our digital attack scenario, we chose to position ${PS}_{1}$ and ${PS}_{2}$ along the top and bottom edges of the image, a similar strategy can be applied to position them along the top and bottom edges or the left and right edges in the context of physical attacks. This choice is grounded in the principle that placing ${PS}_{1}$ and ${PS}_{2}$ in either of these two position configurations yields analogous and consistent outcomes.

\textbf{Color $C$:} In our investigation, we employ color to encapsulate the visual attributes of the camera patch, with its representation denoted as $C (r, g, b)$, where $r$, $g$, and $b$ signify the red, green, and blue channel values of the color, respectively. Within the context of our digital attack experiments, our simulation-based approach affords us the flexibility to explore a range of color camera patch effects, rendering color choice unconstrained in this domain. However, when implementing the physical attack, owing to practical limitations, we exclusively opt for three distinct color camera patches, pink, blue, and green, to execute the physical attacks. This selection is guided by considerations of experimental feasibility. Importantly, our subsequent ablation experiments underscore that the attack success rate of ADCP exhibits minimal correlation with color. Hence, the selection of these three specific colors for our physical experiments is justified and aligned with the findings of our study.

\textbf{Width $W$:} To effectively encapsulate the scale of the camera patch in the horizontal direction, we introduce the parameter width ($W$). As delineated in Figure \ref{figure3}, the width $W$ is designed to measure the horizontal extent of the camera patch. In prescribing the value of this parameter, we align it with the width of the image to ensure its adaptability across diverse image dimensions. Specifically, we define the width $W$ within a range spanning from 0.1 to 0.9, with intervals set at 0.2. The chosen value of $W$ signifies the proportion of the camera patch's width to that of the overall image.

\textbf{Transparency $TS$:} To further encapsulate the attributes of camera patches, we introduce the parameter transparency ($TS$). Transparency $TS$ encapsulates the extent to which the camera patch exhibits transparency within the image. We restrict the definition of $TS$ within the confines of 0.1 to 0.9, with intervals set at 0.1. The transparency parameter $TS$, in a sense, embodies the perceptual visibility of the camera patch. A diminished $TS$ value implies heightened transparency of the camera patch within the image, thereby rendering it less discernible. Conversely, an elevated $TS$ value accentuates the presence of the camera patch, thereby amplifying the likelihood of capturing the observer's attention.

\subsection{Camera patch attack}

We adopt the vector $\theta=\{PS,C,W,TS\}$ to succinctly represent the camera patch generated in our approach. The size of $\theta$ adheres to the pre-set boundaries defined by the vectors ${\theta}_{min}$ and ${\theta}_{max}$, where ${\theta}_{min}$ and ${\theta}_{max}$ can be adjusted. Hence, the process of crafting adversarial examples within a digital context by employing predetermined physical parameters can be elucidated as follows:

\begin{equation}
    \label{Formula 2}
    {X}_{adv} = S(X, \theta) \quad \theta \in ({\vartheta}_{min},{\vartheta}_{max})
\end{equation}
Where, $S$ represents a simple linear fusion of the generated simulation camera patch and the clean sample to obtain the digital adversarial sample ${X}_{adv}$.

\begin{figure}
\centering
\includegraphics[width=1\columnwidth]{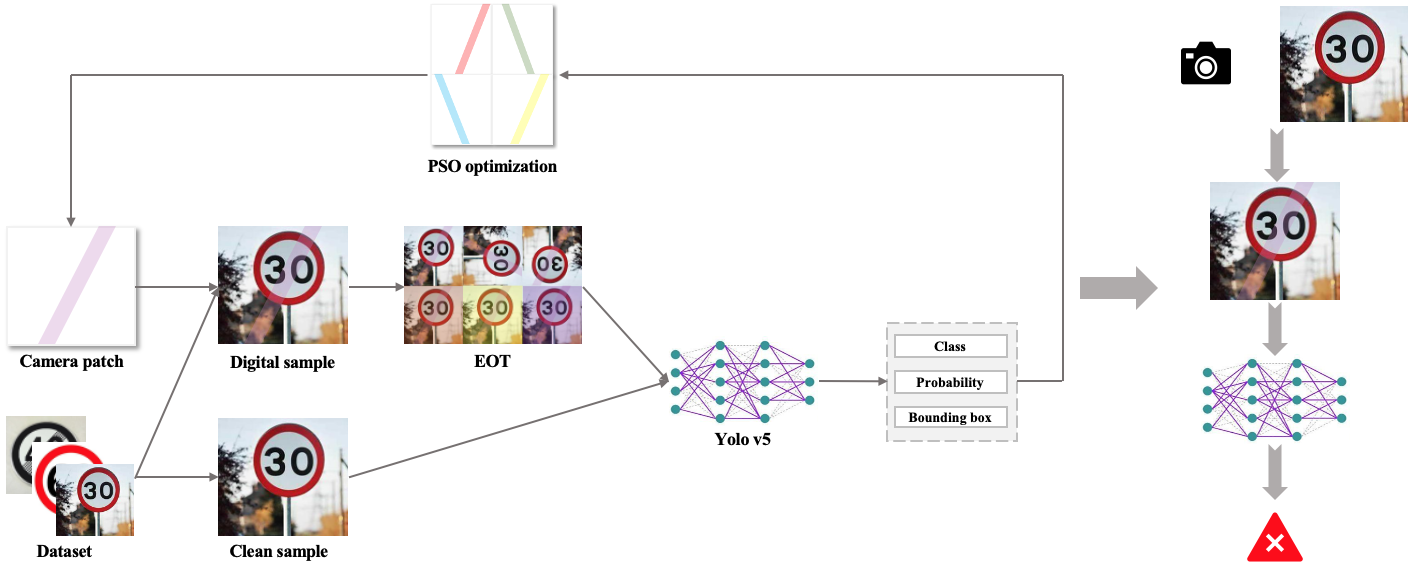} 
\caption{Overview of ADCP attack. The left side shows the optimization process of our method in a digital environment, and the right side indicates that our method generates physical samples in a physical environment. After the robust digital adversarial samples are optimized in the digital environment, perturbations will be deployed in the physical world to generate physical samples.}.
\label{figure4}
\end{figure}

Figure \ref{figure4} provides an illustration of our attack strategy. Within our study, in the digital realm, we simulate camera patches with an expansive spectrum of colors to execute the attack. Meanwhile, in the physical realm, we opt for fixed-color camera patches (pink, green, blue) for the attack. In an endeavor to surmount the inherent experimental disparity between simulated and physical samples, we introduce the concept of Expectation Over Transformation (EOT) \cite{ref37}. EOT stands as a well-established universal framework designed for the generation of robust physical adversarial examples. In the context of EOT, a transformation $\mathcal{T}$ is presented, encapsulating the distribution of all possible transformations. EOT acts as the conduit through which we simulate the transition from the digital domain to the physical realm. In the context of 2D adversarial examples, the transformation set $\mathcal{T}$ comprises rotation, variance, dimming, Gaussian noise, image translation, and other pertinent alterations. By incorporating these different transformations, we optimize to generate adversarial examples that can remain adversarial on transformation $\mathcal{T}$. Consequently, the efficacy of physical attacks is bolstered, thereby attenuating the influence of experimental deviations. By virtue of the incorporation of EOT, we successfully mitigate the impact of experimental discrepancies on physical attacks. Consequently, the ultimate representation of the physical sample is encapsulated as follows:

\begin{equation}
    \label{Formula 4}
    {X}_{adv} = {\mathbb{E}}_{t \sim \mathcal{T}}(t(S(X, \theta),\theta)) \quad \theta \in ({\vartheta}_{min},{\vartheta}_{max})
\end{equation}

Within the scope of this study, our primary objective centers around the simulation and optimization of the physical parameter $\theta$, a precursor to generating the most potent adversarial camera patch. The pivotal criterion is to produce an adversarial sample ${X}_{adv}$, engendered by the physical parameter, that yields successful deception of the object detector. As a consequence, the object detector would either be unable to correctly identify the object or be prompted to misidentify it. A distinguishing aspect of this study is its departure from preceding camera-based endeavors \cite{ref7,ref8}, where the aim is to enhance methodological realism. Specifically, this work operates within the framework of a black-box attack scenario, signifying that access to intricate details such as the target model's network structure is precluded. Instead, only the target class output (${V}_{cls}$) and its associated confidence score (${V}_{obj}$) are accessible. This context precipitates the formulation of a novel approach, wherein ${V}_{obj}$ is harnessed as the adversarial loss. In light of this, the objective is framed as the optimization of the physical parameters characterizing the camera patch, with the intent of minimizing the adversarial loss. This intricate process seeks to generate adversarial samples that successfully fool the object detector. The crux of this optimization endeavor can be succinctly encapsulated as follows:

\begin{equation}
    \label{Formula 5}
    \mathop{\arg\min}_{\theta}{\mathbb{E}}_{t \sim \mathcal{T}}({V}_{obj} \leftarrow t(f({X}_{adv}),\theta))
\end{equation}

To attain the pinnacle of global optimization, we deploy the efficacious Particle Swarm Optimization \cite{ref9} algorithm, aimed at propelling rapid algorithmic convergence and acquiring the optimal solution. PSO constitutes an evolutionary algorithm, with its inception rooted in the intricate behavioral patterns of bird predation. This algorithm is adeptly harnessed for the resolution of optimization conundrums. In the realm of PSO, each optimization quandary's solution embodies a metaphorical "bird" within the expansive search space, a construct referred to as a "particle". These particles collectively manifest distinct fitness values, meticulously dictated by the targeted optimization function. Furthermore, each particle boasts a velocity vector that orchestrates the magnitude and direction of its navigational trajectory. Within this intricate framework, all particles harmoniously trail the trajectory of the preeminent particle, christened the "best particle", as they adroitly explore the expansive solution space in unison.

The PSO algorithm commences with the initialization of an assembly of random particles, essentially, a suite of stochastic solutions. Subsequently, the algorithm embarks upon iterative cycles, meticulously navigating the intricacies of the solution space to uncover the paramount solution. Within each iteration, every particle undergoes a self-update procedure, guided by the pursuit of two pivotal extremal values. Primarily, the particle endeavors to ascertain the optimal solution exclusive to its individual trajectory. This optimal solution, colloquially termed the "individual extrema", is denoted as ${P}_{best}$. In addition, the particle remains attuned to the best solution identified within the entire particle population, a global pinnacle of attainment labeled as ${G}_{best}$. Once both extremal values are discerned, the particle undertakes a momentous transformation of its velocity and position. These modifications are executed with a meticulous formula, devised to optimize the particle's traversal through the solution space while aligning with the core principles of PSO's iterative dynamics. The velocity and position of the particles are updated according to the following formula:

\begin{equation}
    \label{Formula 11}
    {v}_{i}^{j+1}=\omega{v}_{i}^{j}+{c}_{1}{r}_{1}({\theta}_{i,best}^{j}-{\theta}_{i}^{j})+{c}_{2}{r}_{2}({\theta}_{best}^{j}-{\theta}_{i}^{j})
\end{equation}

\begin{equation}
    \label{Formula 12}
    {\theta}_{i}^{j+1}={\theta}_{i}^{j}+{v}_{i}^{j+1}
\end{equation}

where $i \in (1,k)$, $k$ denotes the number of populations. $j$ denotes the current iteration number and $\omega$, ${c}_{1}$, ${r}_{1}$, ${c}_{2}$, ${r}_{2}$ denote the hyperparameters of the PSO algorithm. $v$ is the velocity of the particle and $\theta$ is the position of the particle (i.e., the current solution).

Algorithm \ref{Algorithm 1} outlines the process of the proposed ADCP employing PSO for optimization. ADCP takes several inputs: a clean sample $X$, the target detector $f$, ground truth label $Y$, maximum number of iterations $Step$, and PSO hyperparameters $\omega$, ${c}_{1}$, ${r}_{1}$, ${c}_{2}$ and ${r}_{2}$ which can be determined by the attacker. The pseudocode details are elucidated in Algorithm \ref{Algorithm 1}. Initially, it initializes the initial velocity and position of each particle within the swarm. Subsequently, in each iteration, the camera patch represented by each particle is combined with the clean sample to generate an adversarial sample. The confidence score of the particle, which serves as its fitness value, is then obtained. It's worth noting that a lower confidence score indicates a higher fitness level for that particle. For each particle, if its corresponding adversarial sample induces the model to misidentify or fail to recognize the target, the position information of the particle is recorded as the required physical parameters for the camera patch. Ultimately, in every iteration, the historical optimal solution ${P}_{best}$ for each particle and the historical optimal solution ${G}_{best}$ for the entire swarm are determined. These two optimal solutions are employed to update the velocity and position information of each particle. The program ultimately outputs the physical parameter $\theta$ of the camera patch, which is utilized for subsequent physical attacks.

\begin{algorithm}[t]
	\renewcommand{\algorithmicrequire}{\textbf{Input:}}
	\renewcommand{\algorithmicensure}{\textbf{Output:}}
	\caption{Pseudocode of ADCP}
	\label{Algorithm 1}
	\begin{algorithmic}[1]
	
		\REQUIRE Input $X$, Object detector $f$, Ground truth label $Y$, Max step $Step$, $\omega$, ${c}_{1}$, ${r}_{1}$, ${c}_{2}$, ${r}_{2}$;
		\ENSURE A vector of parameters $\theta$;

        \FOR{each particle $i$}
        \STATE Initialize position ${\theta}_{i}$ randomly;\\
        \STATE Initialize velocity  ${v}_{i}$ randomly;
        \ENDFOR

        \FOR{$j$ $\leftarrow$ 0 to $Step$}
        \FOR{each particle $i$}
            \STATE ${X}_{i}^{j}=S(X,{\theta}_{i}^{j}(PS,C,W,TS))$;
            \STATE $[{V}_{pos},{V}_{obj},{V}_{cls}] \leftarrow f({X}_{i}^{j})$;
            \STATE Obtain the individual optimal value ${P}_{i,best}^{j}$;
            \STATE Obtain the global optimal value ${G}_{best}^{j}$;
            \IF{$f({X}_{i}^{j}) \rightarrow \varnothing $ or $ {V}_{cls} \neq Y$}
                \STATE Output $\theta = {\theta}_{i}^{j}(PS,C,W,TS)$;
                \STATE Exit();
            \ENDIF
        \ENDFOR
        
        \FOR{each particle $i$}
        \STATE ${v}_{i}^{j+1}=\omega{v}_{i}^{j}+{c}_{1}{r}_{1}({\theta}_{i,best}^{j}-{\theta}_{i}^{j})+{c}_{2}{r}_{2}({G}_{best}^{j}-{\theta}_{i}^{j})$;
        \STATE ${\theta}_{i}^{j+1}={\theta}_{i}^{j}+{v}_{i}^{j+1}$;
        \ENDFOR
        \ENDFOR		
		
	\end{algorithmic}  
\end{algorithm}

\section{Evaluation}

\subsection{Experimental setting}

\textbf{Dataset:} In this study, the TT100K dataset is chosen as the fundamental source for both model training and attack experiments. Originating from the collaborative efforts of the Tsinghua-Tencent joint laboratory, the TT100K dataset endeavors to construct an extensive Chinese traffic sign benchmark. This repository encompasses over 30,000 traffic sign instances extracted from approximately 100,000 images, characterized by a resolution of $2048 \times 2048$ pixels and a diverse array of lighting and meteorological conditions. Given the specific focus of our research on physical attacks targeting traffic sign detectors, we meticulously curate the original dataset to retain solely those traffic signs with no fewer than 100 samples. This process ensures the training of a robust traffic sign detection model. Consequently, a new dataset emerges, denominated as TT100K-CameraPatch (TT100K-CP), encompassing a total of 10,427 images. Throughout our experiments, we partition the TT100K-CP dataset into training, testing, and validation sets, comprising 7,568, 1,889, and 970 images, respectively. Moreover, for the digital attack experiment, the segregated test set is designated as our attack dataset, enabling a comprehensive evaluation and validation of our proposed method.

\textbf{Object detector:} In this study, we have opted for YOLO v5 \cite{yolov5} as the object detector to train the traffic sign detection model for Chinese road signs. The selection of YOLO v5 is based on its dual attributes of speed and efficiency, along with its widespread adoption within the field. To expedite convergence, we have utilized pre-trained weights from YOLO v5 and subsequently fine-tuned the model using the TT100K-CP dataset. This approach not only accelerates the training process but also aids in model adaptation to the specific dataset. By following this approach, we have achieved significant results on the test set, attaining an impressive average accuracy of up to 80\%.

\textbf{Experimental devices:} The devices include a support frame, camera patches, and an Redmi K40. Among them, the support frame can be firmly installed on the phone to ensure the stability and repeatability of the experiment. This device configuration allows us to conduct physical attack experiments in a real-world environment to verify the effectiveness of our proposed approach in practical scenarios. In our experiments, we particularly emphasize the versatility of the experimental devices. After verification, we find that different camera models do not affect the effectiveness of our proposed method. This means that our method can maintain a certain degree of robustness and feasibility on different camera devices. This result strengthens the practical application value of our study, making our method potentially applicable to a variety of camera devices.

\textbf{Evaluation criteria:} The core objective of ADCP is to affix meticulously designed camera patches onto the camera lens, thereby inducing erroneous recognition by the object detector under specific conditions or impeding its accurate identification of authentic road signs. Aligned with this goal, we embrace the Attack Success Rate (ASR) as the pivotal metric to gauge the efficacy of our proposed methodology. The ASR is defined as follows:

\begin{equation}
\label{eq:Positional Encoding}
\begin{split}
    &{\rm ASR}(X) = 1-\frac{1}{N}\sum_{i=1}^{N}F({label}_{i})\\
    &F({label}_{i})=
        \begin{cases}
        1 & {label}_{i} \in {L}_{pre} \\
        0 & otherwise
        \end{cases}
\end{split}
\end{equation}

where $N$ represents the number of true positive samples detected by the target detector $f$ in the data set $D$ in the case of no attack, and ${L}_{pre}$ represents the set of all labels detected in the case of attack. A higher ASR indicates a more effective attack.

\textbf{Baselines:} We select the existing camera-based physical attacks as the baselines, including AdvCS \cite{ref7} and TTP \cite{ref8}.

\textbf{Other details: }We set the hyperparameters of PSO as follows: $\omega=0.9$, ${c}_{1}=1.6$, ${r}_{1}=1$, ${c}_{2}=2.0$, ${r}_{2}=1$. For all attack experiments, we execute on a single NVIDIA GeForce RTX 2080 Ti GPU.

\subsection{Evaluation of effectiveness}

\textbf{Digital test:} In this study, we undertake a comprehensive series of digital experiments to validate the efficacy of the proposed approach within a simulated environment. This endeavor involves a sequence of ablation experiments meticulously designed to thoroughly assess the performance of our method in a digital context. These ablation experiments serve the dual purpose of delineating the attack configuration for the subsequent physical experiments and analyzing the adversarial impact of ADCP under varying configurations. In this study, two pivotal factors, namely the width and transparency of the camera patch, are identified as the primary drivers of influence. It is easy to know that excessively small camera patches might fail to yield robust adversarial effects, while overly wide patches could compromise the subtlety of the attack. Similarly, camera patches with high transparency may not ensure robust adversarial effects, whereas those with low transparency might undermine the cloak of stealthiness. In light of these considerations, a delicate equilibrium is sought between optimizing attack effectiveness and preserving stealth. Specifically, the width parameter of the camera patch is spanned across the range of 0.1 to 0.9 with a step size of 0.2. Similarly, the transparency parameter traverses the range of 0.1 to 0.9 with an increment of 0.1. By conducting ablation experiments across various combinations of width and transparency values within the digital realm, we endeavor to unravel the intricate interplay between these parameters and their impact on attack effectiveness.

\begin{table*}
	\centering
    
    \caption{ Results of ablation experiments.}.
    \label{Table2}
    \resizebox{\textwidth}{!}{

	\begin{tabular}{ccccccccccc}
		\hline
		\multirow{2}*{$TS$} & \multicolumn{2}{c}{$W=0.1$} & \multicolumn{2}{c}{$W=0.3$} & \multicolumn{2}{c}{$W=0.5$} & \multicolumn{2}{c}{$W=0.7$} & \multicolumn{2}{c}{$W=0.9$}\\
		\cmidrule(r){2-3}
        \cmidrule(r){4-5}
        \cmidrule(r){6-7}
        \cmidrule(r){8-9}
        \cmidrule(r){10-11}

		~ & ASR (\%) & Query & ASR (\%) & Query & ASR (\%) & Query & ASR (\%) & Query & ASR (\%) & Query \\
		\hline

        0.1  &40.07&342.86&51.47&279.15&56.62&254.54&55.88&252.63&55.88&238.96\\
        \hline
        0.2 &74.63&180.78&82.35&119.05&86.03&95.29&88.24&82.90&87.13&84.38\\
        \hline
        0.3  &93.75&82.39&96.69&41.27&97.79&27.61&98.53&23.08&98.90&15.62\\
        \hline
        0.4  &96.69&46.17&99.63&15.72&100.00&8.55&100.00&4.25&100.00&4.15\\
        \hline
        0.5  &98.53&35.68&99.26&13.68&100.00&4.72&100.00&2.93&100.00&2.25\\
        \hline
        0.6  &98.53&29.50&100.00&10.78&100.00&3.50&100.00&2.38&100.00&1.64\\
        \hline
        0.7  &98.53&25.11&99.63&10.04&100.00&3.02&100.00&1.88&100.00&1.36\\
        \hline
        0.8  &98.53  & 23.85 & 99.26 & 9.44 & 100.00 & 2.79 & 100.00 & 1.62 & 100.00 & 1.28 \\
        \hline
        0.9  & 98.16 & 25.06 & 99.63 & 6.81 & 100.00 & 2.15 & 100.00 & 1.67 & 100.00 & 1.23 \\
        \hline

	\end{tabular}
}
\end{table*}

\begin{figure}
\centering
\setlength{\abovecaptionskip}{-1pt}
\includegraphics[width=1\columnwidth]{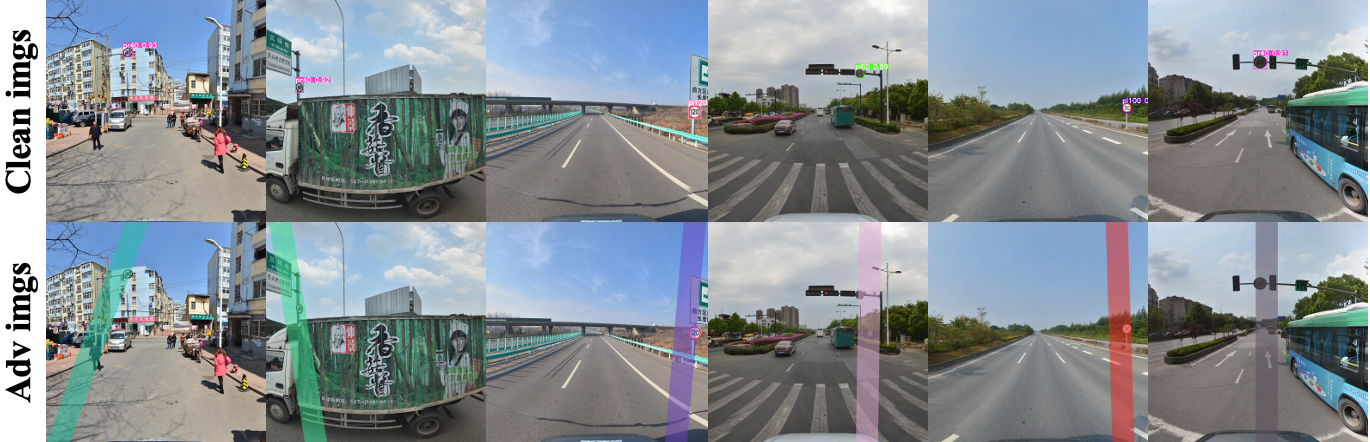} 
\caption{Digital samples.}.
\label{figure6}
\end{figure}

Table \ref{Table2} presents a comprehensive overview of the outcomes derived from the digital ablation experiments. Thorough scrutiny of these experimental findings allows us to formulate the following four pivotal conclusions: Firstly, it is evident that ADCP consistently demonstrates proficient digital attack outcomes across a spectrum of experimental configurations. Remarkably, even in the most challenging scenario ($W=0.1$, $TS=0.1$), ADCP still achieves a commendable attack success rate of 40.07\%. Secondly, there exists a discernible correlation between the augmentation of both the width and transparency of the camera patch and the amplification of ADCP's attack success rate. This alignment with our expectations is mostly upheld, except for the isolated case ($W=0.1$, $TS=0.9$). It is pertinent to mention that, in our experimental context, a higher value of the transparency parameter $TS$ actually signifies a diminished transparency of the camera patch. Thirdly, a pinnacle attack performance is realized by ADCP when the camera patch's width is set to 0.5 and the transparency to 0.4. This signifies that further incremental adjustments in either width or transparency do not significantly elevate the attack success rate, as ADCP has already attained a 100\% success rate in this configuration. Finally, an in-depth examination of the outcomes delineated in Table \ref{Table2} unveils that ADCP attains robust attack performance when the transparency $TS$ reaches 0.3. Based on this insight, we have delimited the configuration range for physical attacks as follows: $0.1 \leq W \leq 0.3$, $0.3 \leq TS \leq 0.5$. Furthermore, in other digital attack experiments, we have consistently set the width to 0.1 and transparency to 0.5. Figure \ref{figure6} graphically presents the adversarial samples engendered by our approach with a width of 0.1 and a transparency of 0.5. The top row showcases the detection outcomes for clean samples, while the subsequent row showcases the detection outcomes for adversarial samples. This visual confirmation substantiates the efficacy and adversarial potency of our methodology. These salient conclusions yield profound insights into our research, fostering an enhanced understanding of ADCP's attack characteristics, and furnishing pragmatic guidance for the configuration of physical attacks in real-world applications. 

\textbf{Physical test:} To conduct a comprehensive and rigorous assessment of the efficacy of our proposed approach in a physical context, we bifurcated the physical experiments into two distinct phases: indoor testing and outdoor testing. This division serves a dual purpose: firstly, it ensures that our attack method remains impervious to extraneous environmental influences during indoor testing; secondly, it facilitates the validation of ADCP's robustness within authentic outdoor settings. In the indoor testing phase, we utilize printed road signs as the test medium. Conversely, in the outdoor testing phase, we opt for genuine road signs from the real world as our testing subjects. This strategic selection effectively scrutinizes the real-world applicability and adversarial impact of our method. Given the inherent variability and unpredictability of outdoor environments, this testing phase is particularly adept at simulating the practical performance of our method within genuine usage scenarios.

\begin{figure}
\centering
\setlength{\abovecaptionskip}{-1pt}
\includegraphics[width=1\columnwidth]{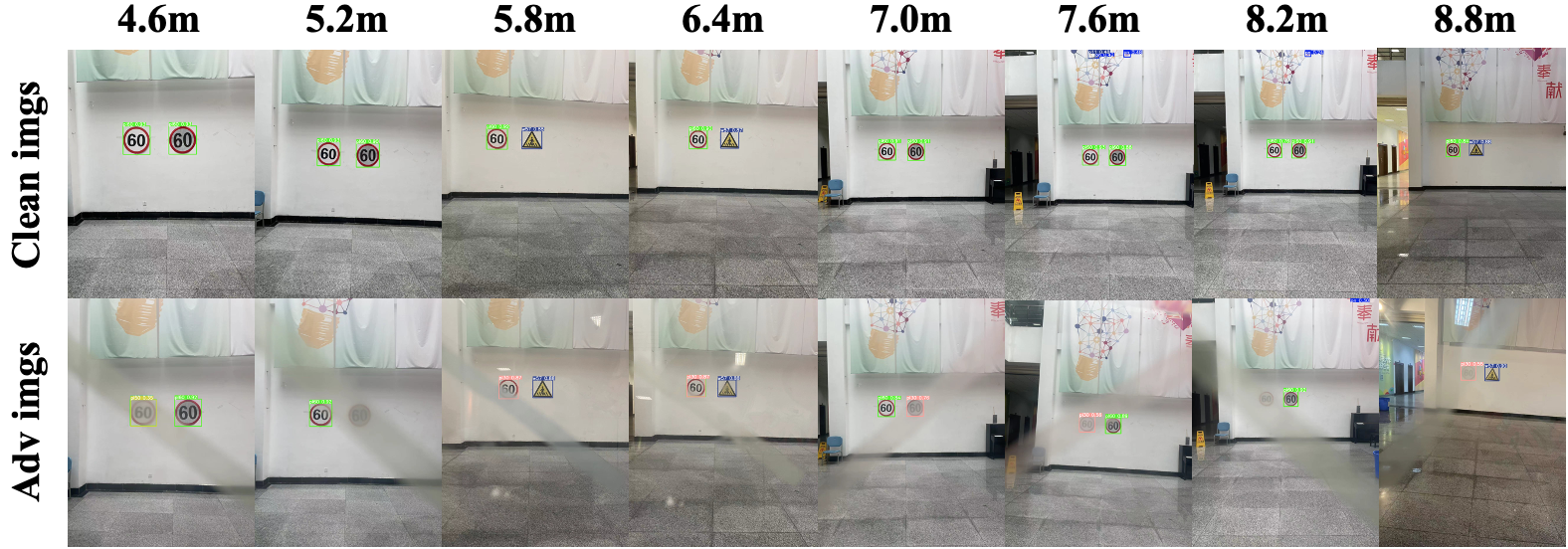} 
\caption{Indoor test.}.
\label{figure7}
\end{figure}

Indoor Testing Phase: To comprehensively assess the adversarial impact of our method across varied conditions, we conduct two distinct sets of experiments during the indoor testing phase. In the initial set, we employ identical road signs, while the subsequent set involved different road signs. These experiments are video-recorded, with the resultant footage being divided into frames to generate physical samples. Subsequently, the Attack Success Rate is calculated based on these samples. Our findings reveal that in the two sets of experiments, we accumulate 1630 and 4772 physical samples, respectively, achieving attack success rates of 78.16\% and 96.31\%. To gain a more profound understanding of ADCP's adversarial effect across different distances, we conduct a distance-based subdivision of the attacks and perform statistical analysis. The results of this analysis are presented in Table \ref{Table3}. These outcomes unequivocally underscore the efficacy and resilience of our method across the various tested distances within the indoor testing context. To present these results in a more intuitive manner, Figure \ref{figure7} showcases the physical samples employed in our indoor testing phase. This visual representation emphasizes that our method effectively executes physical attacks irrespective of the distance involved. During this testing stage, we opted for a camera patch width of $W=0.1$ and a transparency of $TS=0.3$. Of notable importance is the fact that the presence of perturbations remains nearly imperceptible to the human eye without meticulous observation. This underscores the inherent imperceptibility of our attack method. Through this series of indoor testing experiments, our confidence in the method's effectiveness within a controlled environment has been substantially fortified, thus laying a robust foundation for our research.

\begin{table*}
	\centering
    \small
    
    \caption{Results of indoor test experiments (ASR).}.
    \label{Table3}
    \resizebox{\textwidth}{!}{

	\begin{tabular}{ccccccccc}

    \hline
    Distance & 4.6m & 5.2m & 5.8m & 6.4m & 7.0m & 7.6m & 8.2m & 8.8m \\
    \hline
    pl60+pl60&77.11\%&88.03\%&92.89\%&83.07\%&45.45\%&55.96\%&89.19\%&83.00\% \\
    \hline
    pl60+w57&98.02\%&90.12\%&92.13\%&98.58\%&98.53\%&97.82\%&99.50\%&97.57\% \\
    \hline

	\end{tabular}
}
\end{table*}

\begin{figure}
\centering
\setlength{\abovecaptionskip}{-1pt}
\includegraphics[width=1\columnwidth]{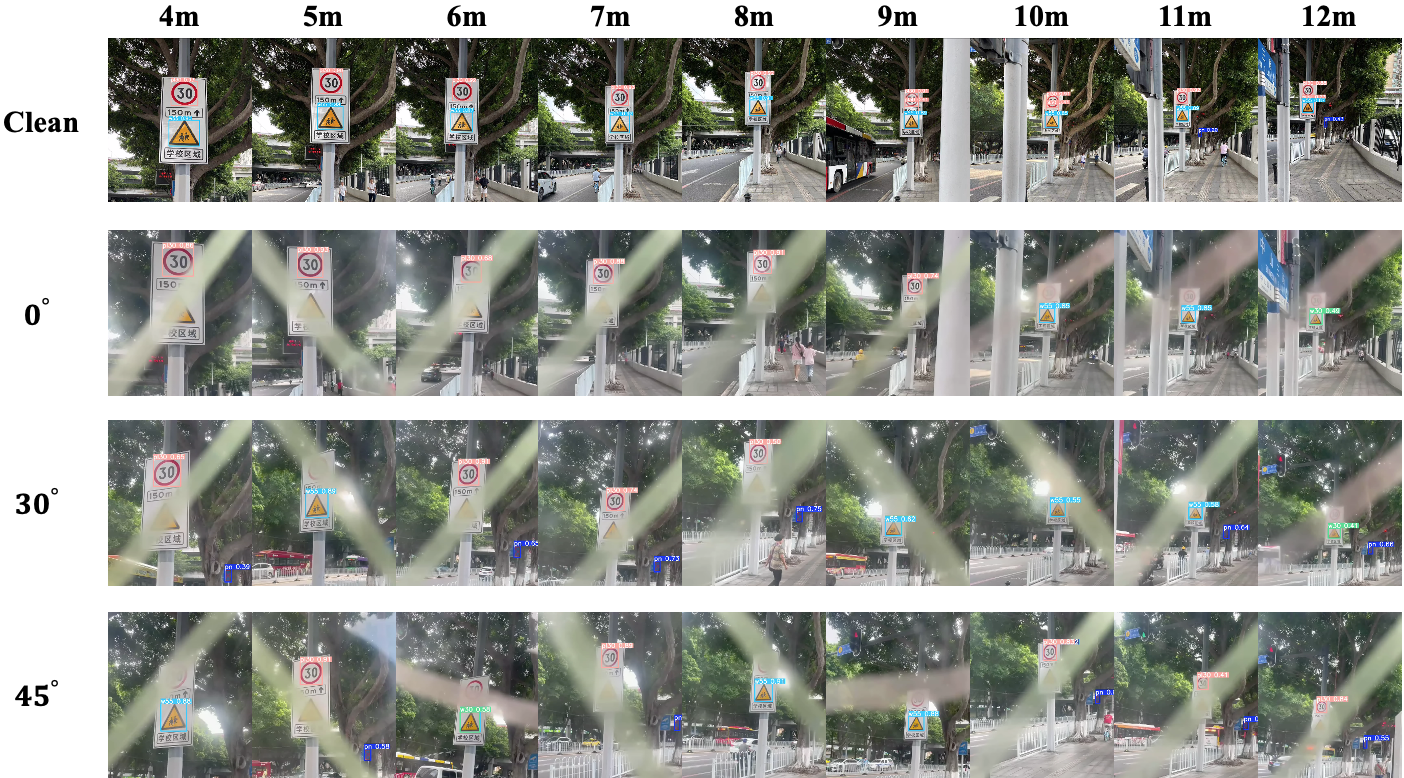} 
\caption{Outdoor test.}.
\label{figure8}
\end{figure}

\begin{table*}
	\centering
    \small
    
    \caption{Experimental results of outdoor testing (ASR).}.
    \label{Table4}
    \resizebox{\textwidth}{!}{

	\begin{tabular}{cccccccccc}

    \hline
    Distance & 4m & 5m & 6m & 7m & 8m & 9m & 10m & 11m & 12m \\
    \hline
    ${0}^{\circ}$ &56.88\%&100.00\%&60.28\%&61.26\%&98.11\%&100.00\%&84.34\%&87.50\%&100.00\% \\
    \hline
    ${30}^{\circ}$ &100.00\%&100.00\%&85.85\%&87.16\%&100.00\%&72.90\%&100.00\%&100.00\%&59.05\% \\
    \hline
    ${45}^{\circ}$ &79.61\%&63.03\%&87.39\%&73.58\%&93.41\%&100.00\%&76.36\%&87.96\%&55.56\% \\
    \hline

	\end{tabular}
}
\end{table*}

Outdoor Testing Phase: In the outdoor testing phase, we transition to utilizing genuine road signs to subject our method to realistic attack scenarios. This shift allows us to more comprehensively assess the method's robustness in real-world conditions. Our objective is to evaluate the method's effectiveness across various distances and angles, mirroring the unpredictability of genuine outdoor environments. During the outdoor testing, we adhere to an attack configuration of $0.1 \leq W \leq 0.3$ and $0.4 \leq TS \leq 0.5$, thereby striking a balance between attack effectiveness and stealthiness. Impressively, our outdoor testing yields a total of 2624 genuine physical samples, culminating in a remarkable attack success rate of 83.31\%. This notable outcome underlines the practical viability and adversarial prowess of our approach within outdoor settings. The detailed experimental findings are elaborated in Table \ref{Table4}, which unequivocally illustrates that our method consistently manifests effective and resilient physical attack outcomes across diverse distances and angles. Remarkably, one third of the distance-angle combinations resulted in a 100\% attack success rate. Even in the worst scenario (7m, ${30}^{\circ}$), we achieve a noteworthy attack success rate of 55.56\%. Figure \ref{figure8} visually captures the adversarial samples generated by ADCP during outdoor testing, further underscoring the method's capacity to execute impactful physical attacks across varying distances and angles. This comprehensive validation affirms the robustness and applicability of our method in complex outdoor scenarios. The series of outdoor experiments has yielded highly favorable outcomes, bolstering our confidence in the practical applicability of our approach within real-world contexts.

Table \ref{Table5} presents a comprehensive comparison of experimental outcomes between our proposed ADCP method and the baseline approach. By scrutinizing these comparative findings, we readily discern that our ADCP method consistently exhibits a more robust adversarial effect in both digital and physical settings when contrasted with the baseline method. This outcome underscores the inherent superiority of our approach. It is imperative to acknowledge that our ADCP method operates in a black-box scenario, aligning closely with real-world practical applications. Moreover, in contrast to the AdvCS and TTP methodologies that employ multiple camera patches, which consequently yield larger experimental errors and inadequate stealthiness, our ADCP approach's utilization of a solitary camera patch yields a higher attack effectiveness coupled with superior stealthiness. Drawing from the comprehensive analysis of Tables \ref{Table1} and \ref{Table5}, we can confidently assert that our proposed ADCP method bears a more formidable realistic threat compared to AdvCS and TTP methods. We emphatically advocate for the recognition and consideration of our ADCP method. Its impressive experimental performance not only attests to its efficacy but also underscores its potential to pose significant challenges to computer vision systems within authentic environments. ADCP, thus, stands as a significant augmentation to the realm of camera-based physical attacks, deserving attention for its multifaceted capabilities.

\begin{table*}
	\centering
    
    \caption{Comparison of experimental results between our approach and baselines.}.
    \label{Table5}

	\begin{tabular}{cccccc}
    
    \hline
    \multirow{2}*{Method} & \multirow{2}*{Scenario} & \multicolumn{2}{c}{Digital attack} & \multicolumn{2}{c}{Physical attack} \\
    \cmidrule(r){3-4}
    \cmidrule(r){5-6}
    ~ & ~ & ASR & Query & Indoor ASR & Outdoor ASR \\
    \hline
    AdvCS \cite{ref7} & White-box&49.60\%& $\varnothing$ & $\varnothing$ & 73.26\% \\
    \hline
    TTP \cite{ref8} & White-box&42.47\%&$\varnothing$ &$\varnothing$ & 42.27\% \\
    \hline
    ADCP (Ours) & Black-box&93.75\%& 83.39\% & 78.16/96.31\% & 88.31\%  \\
    \hline

    \end{tabular}

\end{table*}

\subsection{Ablation study}

\begin{figure}
\centering
\setlength{\abovecaptionskip}{-1pt}
\includegraphics[width=1\columnwidth]{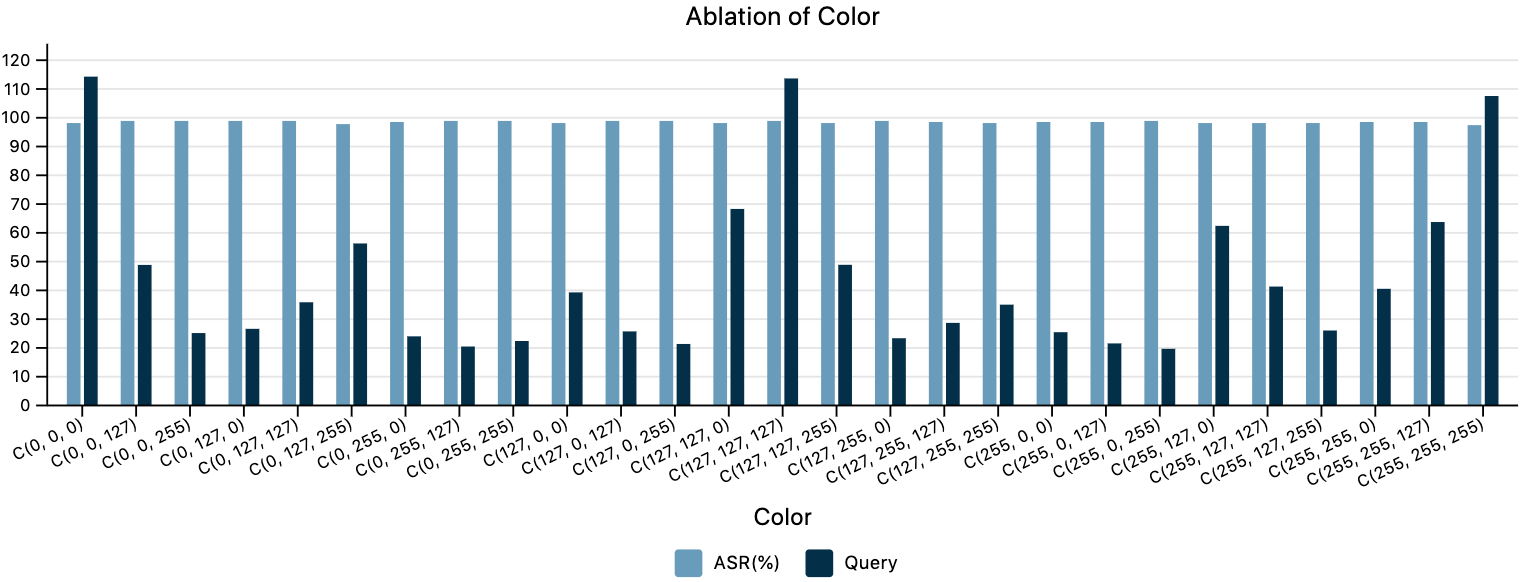} 
\caption{Ablation of $C$.}.
\label{figure9}
\end{figure}

In this section, we delve into an in-depth exploration of yet another pivotal influencing factor: the color of the camera patch. Specifically, our focus encompasses the RGB channel of the color, encapsulating the $r$, $g$, and $b$ values that span the range of (0, 255). Nevertheless, it is evidently impractical to execute ablation experiments for every conceivable color scenario, given the potential explosion in the number of experimental combinations. To strike a balance in our experimental design, we opted for a judicious trade-off. We elect to perform ablation experiments with three distinct values for each of the RGB channels: 0, 127, and 255. This strategic choice enabled us to craft a total of 27 experimental configurations, meticulously crafted to comprehensively illuminate the role of color in the efficacy of ADCP attacks. Figure \ref{figure9} visually encapsulates the outcomes of these color ablation experiments. Of paramount importance is the observation gleaned from these experimental results: color appears to wield limited influence over the adversarial efficacy of ADCP. In essence, camera patches sporting diverse colors seemingly exert relatively minor sway over the potency of the attack. This discovery tangibly bolsters the robustness and universality of our method. The method's ability to maintain a relatively steadfast attack effectiveness across different color variations reinforces its utility and adaptability, thus underscoring its reliability in contexts.

\subsection{Transfer attack}

To glean a more comprehensive understanding of the adversarial prowess of our ADCP methodology, we embark on transfer attack experiments aimed at elucidating its performance against diverse target detectors. In particular, we subject three distinct object detectors to scrutiny: Faster Rcnn \cite{ref39}, Libra Rcnn \cite{ref40}, and RetinaNet \cite{ref41}. To facilitate these assessments, we harness the adversarial examples that successfully attacked YOLO v5 as our attack dataset. In this experimental endeavor, we employ pre-generated adversarial samples and gauge their impact on different object detectors. The chosen object detectors encompass a broad spectrum within the realm of computer vision, encapsulating a variety of detection algorithms and frameworks. By subjecting these detectors to our transfer attack experiments, we unveil insights into the applicability and efficacy of the ADCP method across diverse detector architectures. This analytical exploration allows us to gain a comprehensive grasp of our method's performance, elucidating its potential to successfully challenge a spectrum of target detectors.

Embarking on the domain of digital transfer attack experiments, we pivot our focus toward the utilization of digital samples previously proven effective against YOLO v5. By employing these samples, we orchestrate attacks against alternative object detectors, namely Faster Rcnn, Libra Rcnn, and RetinaNet. To facilitate this analysis, we harness camera patches configured with attack parameters $W=0.1$ and $TS=0.5$. In this digital transfer attack endeavor, our methodology yields notable outcomes: an attack success rate of 80.08\% against Faster Rcnn, 65.04\% against Libra Rcnn, and 80.08\% against RetinaNet. These findings vividly underscore the efficacy of our ADCP approach in orchestrating successful transfer attacks within digital domains. This substantiates our method's versatility and resilience, solidifying its stature as a contender for challenging various object detectors in a digital context.

Transitioning into the domain of physical transfer attack experiments, we select the physical samples that can successfully attack Yolo v5 as the dataset, and perform transfer attacks on these samples against Faster Rcnn, Libra Rcnn and RetinaNet. This experiment aims to verify the transfer attack effect of our ADCP method in a real physical environment. The results of the indoor attack experiments are shown in Table \ref{Table6}. When attacking a combination of road signs pl60+pl60, our method achieves 100\% physical transfer ASR against all detectors. ADCP also achieves robust physical transfer attack when attacking the road sign combination of pl60+w57. Even the worst case, attacking RetinaNet at 8.2 meters, achieves 61.70\% ASR. In the outdoor test, ADCP achieved 100\% attack success rate for the transfer attack on each model in the physical test, both at different distances and angles. This result shows that our method exhibits excellent robustness and reliability against physical transfer attacks. After further analysis, we believe that this is mainly due to the adversarial samples generated by the attack configurations we selected in the outdoor tests ($0.1 \leq W \leq 0.3$, $0.4 \leq TS \leq 0.5$). The perturbation degree of these samples is relatively large, so they can still maintain a high attack effect when they are transferred to other target detectors.

\begin{table*}
	\centering
    
    \caption{ Indoor physical transferability of ADCP.}.
    \label{Table6}
    \resizebox{\textwidth}{!}{
	\begin{tabular}{cccccccccc}
    
    \hline
    ~&Model&4.6m&5.2m&5.8m&6.4m&7.0m&7.6m&8.2m&8.8m  \\

    \hline
    \multirow{3}*{pl60+pl60}& Faster Rcnn&100\%&100\%&100\%&100\%&100\%&100\%&100\%&100\% \\
    \cmidrule(r){2-10}
    ~ & Libra Rcnn&100\%&100\%&100\%&100\%&100\%&100\%&100\%&100\% \\
    \cmidrule(r){2-10}
    ~ & RetinaNet&100\%&100\%&100\%&100\%&100\%&100\%&100\%&100\% \\
    \hline

    \multirow{3}*{pl60+w57}& Faster Rcnn&99.21\%&98.51\%&100\%&99.92\%&96.23\%&95.52\%&93.62\%&96.06\% \\
    \cmidrule(r){2-10}
    ~ & Libra Rcnn&98.41\%&98.88\%&100\%&99.82\%&96.86\%&97.76\%&94.22\%&97.13\% \\
    \cmidrule(r){2-10}
    ~ & RetinaNet&99.21\%&99.26\%&99.79\%&98.92\%&96.23\%&93.28\%&61.70\%&78.14\% \\
    \hline

	\end{tabular}
 }
\end{table*}

\subsection{Deploying ADCP to attack object detectors across various tasks}

In order to assess the universality of ADCP, we deploy it to attack object detectors trained on the COCO dataset \cite{ref42}, selecting Faster Rcnn \cite{ref39} and RetinaNet \cite{ref41} as our test models. The validation set of the COCO dataset is chosen as the attack dataset to evaluate the attack success rate of ADCP. In this experiment, we configure $TS$ to be 0.5 and vary the width parameter ($W$) from 0.1 to 0.9 in increments of 0.2. The experimental results are presented in Table \ref{Table9}, demonstrating that our method effectively launches adversarial attacks on object detectors trained on the COCO dataset. Generally, the attack success rate increases with higher values of the width parameter. In summary, our approach, ADCP, demonstrates the capability to effectively perform attacks against object detectors across various tasks.

\begin{table*}
	\centering
    
    \caption{ Results of ablation experiments.}.
    \label{Table9}
    \resizebox{\textwidth}{!}{

	\begin{tabular}{ccccccccccc}
		\hline
		\multirow{2}*{Models} & \multicolumn{2}{c}{$W=0.1$} & \multicolumn{2}{c}{$W=0.3$} & \multicolumn{2}{c}{$W=0.5$} & \multicolumn{2}{c}{$W=0.7$} & \multicolumn{2}{c}{$W=0.9$}\\
		\cmidrule(r){2-3}
        \cmidrule(r){4-5}
        \cmidrule(r){6-7}
        \cmidrule(r){8-9}
        \cmidrule(r){10-11}

		~ & ASR (\%) & Query & ASR (\%) & Query & ASR (\%) & Query & ASR (\%) & Query & ASR (\%) & Query \\
		\hline

        Faster Rcnn&33.60&365.18&40.21&336.48&42.86&325.17&45.77&311.83&50.79&289.66\\
        \hline
        RetinaNet&44.35&325.50&50.38&292.09&53.82&267.25&56.30&255.44&55.87&252.19\\
        \hline

	\end{tabular}
}
\end{table*}

\section{Conclusion}

In this study, we introduce a pioneering camera-based physical attack method called ADCP, distinguished by its novel black-box approach. Leveraging the Particle Swarm Optimization algorithm, we optimize the physical parameters of the camera patch, yielding a potent adversarial tool for robust physical attacks. A key advantage of ADCP over traditional methods lies in its capacity to achieve attack objectives without necessitating modifications to the target object itself. This innovative design inherently enhances stealthiness, offering a distinctive edge over existing camera-based attack techniques.

Furthermore, ADCP demonstrates superiority in terms of operational simplicity and robustness, utilizing only one camera patch to achieve more reliable and resilient black-box attacks. Our meticulously planned experimental framework and results substantiate the efficacy and resilience of ADCP in real-world applications. Efficacy-wise, ADCP exudes substantial adversarial impact across both digital and physical environments, firmly validating its practical viability. Regarding robustness, our method's impressive attack success rates in physical tests, coupled with its heightened ASR across various scenarios than the baseline method, underscore the method's robustness. Invisibility is another hallmark of our approach, as our camera-based strategy circumvents the need for direct modifications to the target object.

Collectively, our research underscores the security threats posed by ADCP in the physical domain. This detailed exploration of our method's underlying mechanisms holds tangible practical implications, promising to enhance comprehension and application of physical attacks. As such, we advocate for a heightened awareness of the potential risks posed by ADCP in real-world contexts, positioning it as a compelling complement to the camera-based physical attacks.





















\bibliographystyle{elsarticle-harv} 
\bibliography{A}





\end{document}